\documentclass[letterpaper]{article} 
\usepackage{aaai25}  
\usepackage{times}  
\usepackage{helvet}  
\usepackage{courier}  
\usepackage[hyphens]{url}  
\usepackage{graphicx} 
\usepackage{natbib}  
\usepackage{caption} 
\usepackage{algorithm}

\usepackage{bibentry}
\usepackage[table,dvipsnames]{xcolor}
\usepackage{subfigure}  
\usepackage{algpseudocode}

\usepackage{latexsym}
\usepackage{amssymb}
\usepackage{amsmath}
\usepackage{amsthm}
\usepackage{booktabs}
\usepackage{enumitem}
\usepackage{graphicx}
\usepackage{color}
\usepackage{overpic}
\usepackage{multirow}
\usepackage{colortbl}
\usepackage{bbding}
\usepackage{pifont}
\usepackage{fontawesome}

\definecolor{lightgrey}{RGB}{244,244,244}
\definecolor{grey}{RGB}{128,128,128}
\definecolor{midgrey}{RGB}{225,225,225}
\definecolor{forestgreen}{RGB}{47, 159, 87}
\newcommand{\cmark}{\color{forestgreen}\ding{51}}%
\newcommand{\xmark}{\color{red}\ding{55}}%

\usepackage[capitalize]{cleveref}
\crefname{section}{Sec.}{Secs.}
\Crefname{section}{Section}{Sections}
\crefname{table}{Table}{Tables}
\Crefname{table}{Table}{Tables}

\usepackage{soul}
\graphicspath{{figures/}}

\definecolor{Gray}{gray}{0.95}
\definecolor{Cyan}{rgb}{0.88,1,1}

\usepackage{newfloat}
\usepackage{listings}
\DeclareCaptionStyle{ruled}{labelfont=normalfont,labelsep=colon,strut=off} 
\floatstyle{ruled}
\newfloat{listing}{tb}{lst}{}
\floatname{listing}{Listing}
\nocopyright

\title{Frame Order Matters: A Temporal Sequence-Aware Model \\ for Few-Shot Action Recognition}
\author {
    Bozheng Li\thanks{Equal Contribution.}, Mushui Liu$^{*}$, Gaoang Wang, Yunlong Yu\thanks{Corresponding Author}
}
\affiliations{
    Zhejiang University
}

\usepackage{bibentry}
\usepackage[dvipsnames]{xcolor}
\usepackage{subfigure}  
\usepackage{algpseudocode}

\begin{document}

\maketitle

\begin{abstract}
In this paper, we propose a novel \textbf{T}emporal \textbf{S}equence-\textbf{A}ware \textbf{M}odel (TSAM) for few-shot action recognition (FSAR), which incorporates a sequential perceiver adapter into the pre-training framework, to integrate both the spatial information and the sequential temporal dynamics into the feature embeddings. Different from the existing fine-tuning approaches that capture temporal information by exploring the relationships among all the frames, our perceiver-based adapter recurrently captures the sequential dynamics alongside the timeline, which could perceive the order change. To obtain the discriminative representations for each class, we extend a textual corpus for each class derived from the large language models (LLMs) and enrich the visual prototypes by integrating the contextual semantic information. Besides, We introduce an unbalanced optimal transport strategy for feature matching that mitigates the impact of class-unrelated features, thereby facilitating more effective decision-making. Experimental results on five FSAR datasets demonstrate that our method set a new benchmark, beating the second-best competitors with large margins. 
\end{abstract}

\section{Introduction} \label{sec:intro}

Video action recognition~\cite{arnab2021vivit, yang2023aim}, a cornerstone task in video understanding, demands comprehensive understanding and representation of spatial-temporal information. However, the inherent challenges of large data volume and storage complexities of video data often lead to a scarcity of training samples, hindering model development. Few-shot action recognition (FSAR)~\cite{clipfsar, cao2020few, perrett2021temporal, xing2023revisiting} emerges as a promising direction to alleviate this data bottleneck while exploring efficient video encoding architectures. By leveraging a limited number of labeled examples, FSAR aims to enable the rapid recognition of novel action categories.

Learning-based methods have significantly advanced the boundaries of FSAR. Existing approaches have explored diverse avenues, with some focusing on the design of powerful video encoders \cite{thatipelli2022spatio, zhang2021learning} and others concentrating on the development of effective FSL paradigms~\cite{zheng2022few, wu2022motion, perrett2021temporal}. The advent of the Contrastive Language-Image Pre-Training (CLIP)~\cite{clip} has ushered in a more robust visual backbone that supports multi-modal information integration. Consequently, leveraging the pre-trained CLIP image encoder for video representation learning has demonstrated promising potential in various works~\cite{wasim2023vita,ni2022expanding,liu2024omniclip}.

\begin{figure}[t]
   \begin{overpic}[width=\linewidth]{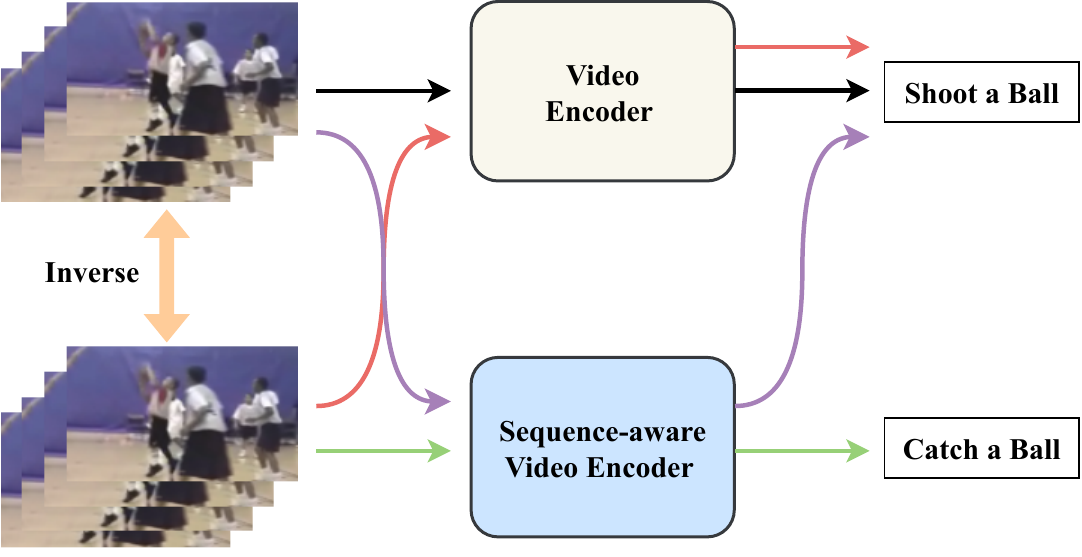} 
   \put(70,48){ \xmark }
   \put(70,40){ \cmark}
   \put(70,20){ \cmark}
   \put(70,10){ \cmark}
   \end{overpic}
   \caption{Illustration of the semantic changing when rearranging the video frames. The existing model could barely distinguish reversed video due to equally treated frame features while our temporal sequence-aware model could capture the differences between the videos with different frame orders. }
   \label{fig:vischange}
\end{figure}
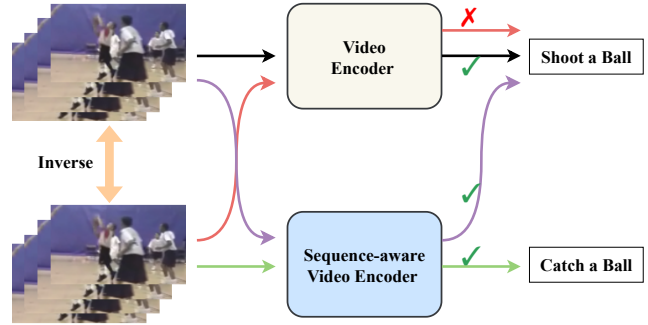

\sloppy
Recent endeavors have investigated the adaptation of CLIP for FSAR tasks. Notably, CLIPFSAR~\cite{clipfsar} pioneered the transfer of the CLIP backbone to FSAR by directly fine-tuning the entire architecture. A series of subsequent works~\cite{pei2023d, maclip, cao2024task} employed Parameter-Efficient Fine-Tuning (PEFT) strategies to transfer the pre-trained backbone to FSAR. These approaches prioritize capturing temporal information by delving into the interrelationships among frames, yet they inadequately consider the temporal sequence. In some cases, the category of a video can be significantly altered by merely rearranging the order of its frames. As demonstrated in \cref{fig:vischange}, reversing the order of input video frames to the models fails in altering the predicted results, while the category semantic has been changed.

Drawing inspiration from the human cognitive process of sequential information processing, we introduce a novel sequence-aware model for FSAR that incorporates a sequential perceiver adapter into the pre-training backbone to integrate both the spatial information and the sequential temporal dynamics into the feature embeddings. Specially, the sequential perceiver adapter captures the inherent order within the video frames, by treating temporal information as a distinct modality alongside spatial features. In each perceiver adapter, a temporal query recurrently compresses temporal information from each frame and injects temporal sequential relation to the pre-trained backbone. To enrich the feature representations of each class, we construct an expanded textual prompt corpus leveraging the Large Language Models (LLMs), which enables discriminative prototype construction with enhanced textual information involved. Besides, we transform the matching of support and query video into an unbalanced optimal transport problem to further improve the matching process.

In summary, the highlights are as follows:
\begin{itemize}
    \item We present a novel temporal sequence-aware model (TSAM) for FSAR that integrates a sequential perceiver adapter into the pre-trained visual backbone to extract both dynamic temporal sequences and spatial information for feature embeddings. The Perceiver Adapter is designed to recurrently integrate and compress spatial information into temporal cues, which could perceive the frame order change.

    \item To enrich the feature representations of each class, we extend a textual corpus derived from LLMs and integrate the textual semantics into the visual prototypes. Besides, we introduce an unbalanced optimal transport strategy for feature matching, which diminishes the influence of non-discriminative, class-irrelevant features.
    
    \item Experimental evaluations conducted on five diverse FSAR datasets have conclusively shown that our proposed TSAM has established a new state-of-the-art benchmark, surpassing the second-best competitor by substantial margins, underscoring its effectiveness and superiority.
    
\end{itemize}
\section{Related Works} \label{sec:related}
\textbf{Few-shot Action Recognition.}
Few-shot action recognition focuses on temporal information modeling and constructing relationships between query and support sets. Previous methods have leveraged metric learning to achieve alignment between query and support sets. 
OTAM \cite{cao2020few}, TRX \cite{perrett2021temporal}, CMOT \cite{lu2021few}, and HyRSM~\cite{wang2022hybrid} utilize dynamic temporal warping, multi-frame matching, optimal transport, and set matching to calculate the video similarities to build the relationship of videos, respectively.
Additionally, other approaches have focused on spatial information in video frames and the fusion of multi-modal information.
MTFAN \cite{MTFAN} learns global motion patterns with a motion modulator.
STRM \cite{STRM} enriches local patch features for spatial-temporal modeling.
MoLo \cite{molo} explicitly extracts motion dynamics within a unified network. Recently, based on the powerful pre-trained model, \textit{e.g.}, CLIP \cite{clip}, CLITSAM \cite{clipfsar} and MA-CLIP \cite{maclip} have shown significant progress by transferring pre-trained image encoders to the video domain and constructing prototypes incorporated with textual information.

\noindent \textbf{Parameter-Efficient Fine-Tuning.}
Parameter-Efficient Fine-Tuning (PEFT) \cite{houlsby2019parameter,VPT} aims to efficiently fine-tune large pre-trained models by adjusting only a few parameters while freezing the backbone, achieving performance similar to full fine-tuning. This paradigm has shown promise in computer vision. Methods like VPT \cite{VPT}, CoOp \cite{coop}, and CLIP-Adapter \cite{clip-adapter} incorporate learnable visual prompts, textual prompts, and adapters to learn discriminative features for downstream tasks. For video recognition, ST-Adapter \cite{pan2022st} uses a 3D-convolution-based adapter to capture temporal information, while AIM \cite{yang2023aim} further reuses the attention layer in the pre-trained model to formulate temporal information. Vita-CLIP \cite{wasim2023vita} introduces learnable prompt tokens in both spatial and temporal dimensions. MA-CLIP\cite{maclip} introduced a textual-guided adapter structure. However, these approaches often overlook the sequential characteristics of video frames and the importance of video frame relationship matching. 

\noindent \textbf{Perceiver} Perceiver~\cite{jaegle2021perceiver} refers to a transformation of asymmetric attention that can distill multi-modal information into a compressed latent bottleneck. Most former works either utilize perceiver to resample high-volume feature tokens~\cite{alayrac2022flamingo} or bridge the feature space of different modality~\cite{acosta2022multimodal}. Recently, Meta utilized perceiver as the video aggregator in LLaMA3.1~\cite{dubey2024llama}, proving its ability in temporal modeling and feature compression for the video domain.

In this paper, we utilize an adapter to transfer the pre-trained backbone for FSAR, emphasizing frame order modeling through the proposed sequential perceiver structure.


\section{Methodology} \label{sec: method}
\begin{figure*}[!htb]
    \centering
    \includegraphics[width=\textwidth]{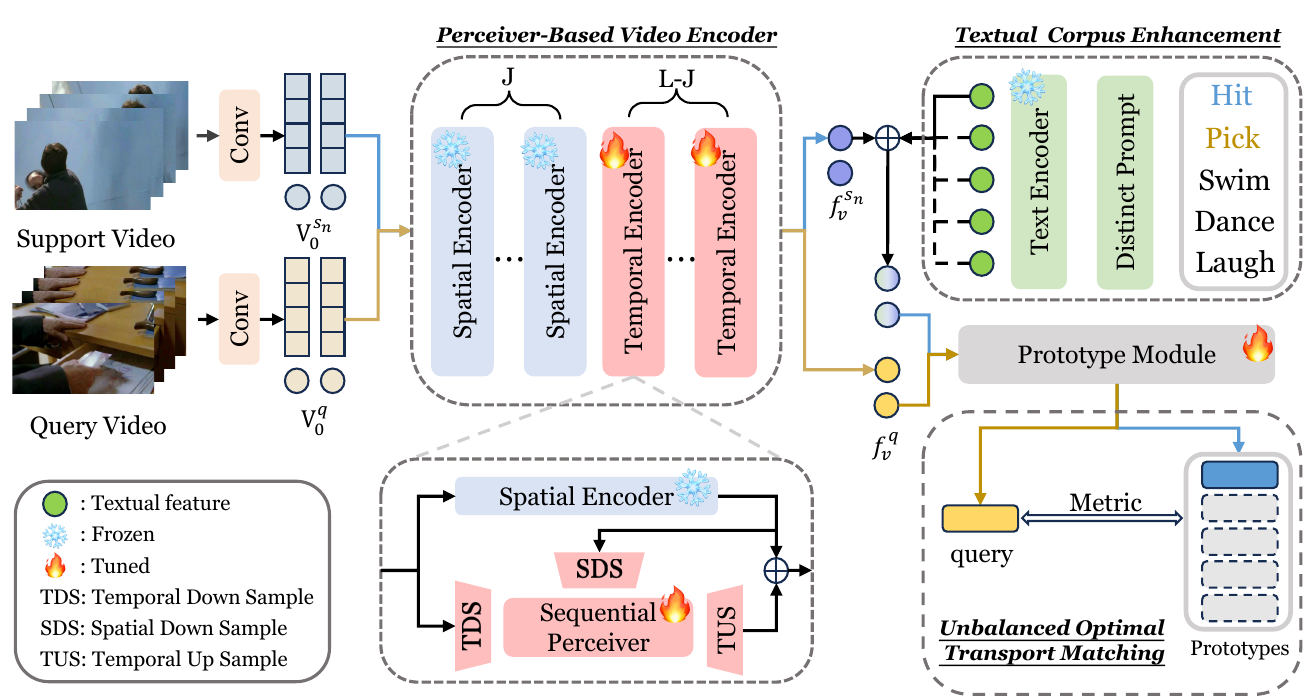}
    \caption{The framework of our proposed temporal sequence-aware model(TSAM), which consists of a perceiver-based video encoder that captures both the temporal sequential dynamics and spatial information, a textual corpus enhancement module that incorporates the class-related semantics into the feature prototype, and an unbalanced optimal transport matching module that enhances the feature matching.    }
    \label{fig:framework}
\end{figure*}

\subsection{Problem Definition}
In FSAR tasks, the video datasets are divided into non-overlapping training $\mathcal{D}_{train}$ and testing $\mathcal{D}_{test}$ splits. Consistent with the prior work \cite{snell2017prototypical}, we employ episodic training by randomly sampling few-shot tasks from the training set. In each task, the video data is divided into a support set $\mathcal{S}$ and a query set $\mathcal{Q}$. The $N$-way $K$-shot task indicates using a support set comprised of $N$ classes, with $K$ samples per class to learn the classifier for the query samples during inference.

\subsection{Temporal Sequence-Aware Model}
\subsubsection{Overall Framework}

Following the sequential nature of video data, we propose a Temporal Sequence-Aware Model (TSAM) framework for FSAR that enriches the frame-level feature representations by incorporating the temporal contexts captured sequentially from the spatial information alongside the temporal dimension. As illustrated in \cref{fig:framework}, the framework consists of a perceiver-based video encoder to extract both the visual and temporal information, a textual corpus enhancement module to enrich the video representations, and an unbalanced optimal transport matching module to relieve the class-unrelated features for decision-making.

Using ViT as the visual backbone, we delve into the details of our proposed framework. Given an input video $ v \in \textstyle \mathbb{R}^{T \times H \times W \times 3} $, where $T$ denotes the uniformly sampled frame numbers, $H$ and $W$ respectively denote the height and width of each frame, we sequentially input the frames into the pre-trained ViT backbone to obtain the feature representations. As the original ViT blocks inherently struggle to capture the temporal dynamics across frames, we propose to extend the block to make it compress spatial information into the temporal cues, and in turn enhance the frame-level feature representations. Assuming a total of $\rm L$ blocks in ViT architecture, we propose to freeze the initial $\rm J$ blocks, preserving their pre-trained weights. For the remaining $\rm L-J$ blocks, we introduce a perceiver adapter in parallel to each block's original spatial encoder. These adapters are then fine-tuned during training, allowing the model to learn temporal cues while leveraging the spatial information encoded by the frozen blocks. To better capture the temporal information, we incorporate a learnable temporal token as an input to the $J+1$-th block. The spatial information encoded with the spatial encoder and the temporal information captured with the perceiver adapter are then fused for the next block. 

To obtain discriminative visual representations for each category with a few visual samples, we propose to incorporate the semantic descriptions related to the categories into the visual presentations, powered with the off-the-shelf large language models (LLMs). Once the feature representations of both support and query samples are obtained, the similarity metric is the key to decision-making. Considering that during the few-shot matching process, no relevant noise frame will harm the matching of the action class, we introduce an unbalanced optimal matching strategy to filter the redundant noise within sampled video frames by punishing many-to-one matching and involving soft condition boundary.

\subsubsection{Perceiver Adapter}
The Perceiver Adapter is a pivotal component within the video encoder, which is designed to capture the temporal information for the feature representations. Perceiver structure was originally introduced in \cite{jaegle2021perceiver} to enhance feature compression and foster cross-modality communication. Inspired by this approach, we view temporal and spatial features as two distinct modalities and introduce a perceiver-based adapter to compress temporal information within spatial features into a temporal query. 

\begin{figure}[t]
   \begin{overpic}[width=\linewidth]{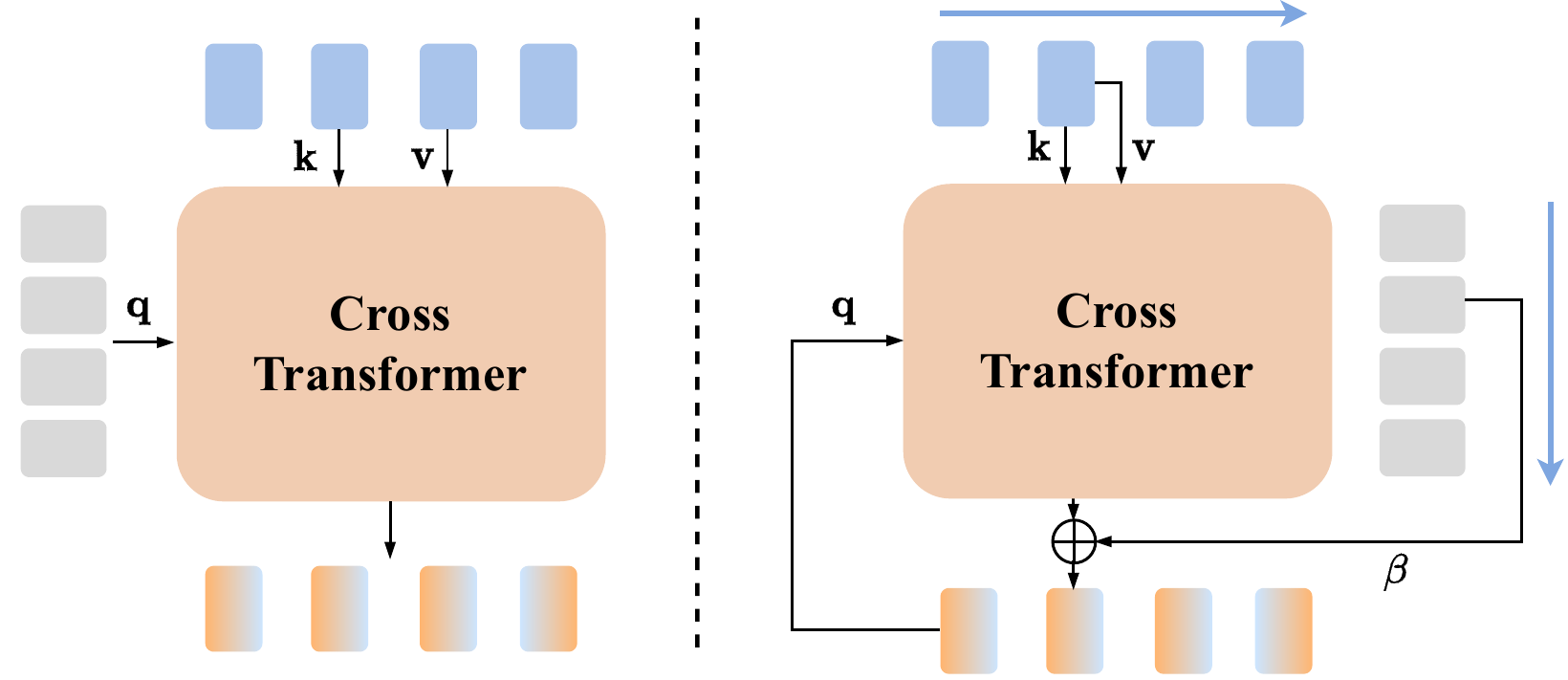} 
   \put(15,-4){\small (a)}
   \put(70,-4){\small (b)}
   \end{overpic}
   \vspace{0.5mm}
   \caption{Illustration of (a) Simple Perceiver-based adapter and (b) our Sequential Perceiver-based adapter.}
   \label{fig:perceiver}
\end{figure}

However, when a straightforward perceiver structure (as shown in \cref{fig:perceiver} (a)) is directly employed as an adapter, two challenges arise. On one hand, utilizing a fully learnable temporal query to condense temporal information within spatial features, while trained on limited data, can inadvertently lead to an overly cohesive prototype space, potentially limiting the model's ability to capture diverse temporal patterns. On the other hand, the straightforward attention mechanisms employed in the perceiver adapter are prone to an inadvertent issue of information leakage \cite{clipfsar, maclip}, where data from subsequent frames inadvertently permeates into preceding ones during processing. This leads to a homogenization effect across temporal dimensions, diminishing the model's capacity to accurately capture and differentiate between temporal dynamics, thereby compromising its overall performance.

To tackle these issues, we propose the sequential perceiver. As shown in \cref{fig:perceiver}(b), instead of learnable query tokens, the sequential perceiver takes the temporal token embedding of the first frame $\mathbf{f}_t^1$ as the query to interact with the spatial feature embedding $\mathbf{f}_s^1$ of the first frame to obtain the temporal embedding $\mathbf{f}_{temp}^1$. Then, the temporal embedding $\mathbf{f}_{temp}^1$ is taken as the query to interact with the spatial feature embedding $\mathbf{f}_s^2$ of the second frame, and so on. Given the feature representations $\mathbf{V}_i\in \mathbb{R}^{T\times U \times D}$ of $i$-th block ($J \leq i \leq L$), the process is first formulated as:
\begin{equation}
\begin{aligned}
    &\mathbf{V}_i^s = {\rm Spatial\_D}({\rm Visual\_B} (\mathbf{V}_i)), \\
    &\mathbf{V}_i^t = {\rm Temporal\_D}(\mathbf{V}_i),
\end{aligned}
\end{equation}
where $\rm Visual\_B$ denotes the visual block, $\rm Spatial\_D$ and $\rm Temporal\_D$ represent the specific modules tasked with diminishing the dimensionality along the spatial and temporal axes, respectively.

Then, for the $k$-th frame, we obtain the temporal feature embedding by:
\begin{equation}
\begin{aligned}
    \mathbf{F}_k^t = {\rm Cross\_Att}(\mathbf{F}_{k-1}^t,~ \mathbf{V}_k^s, ~\mathbf{V}_k^s) + \beta \cdot \mathbf{V}_k^t, 
\end{aligned}
\end{equation}
where $\mathbf{F}_{k-1}^t$ performs as the query and $\beta$ denotes the injection ratio of temporal information. For the first frame, the query is initialized as the temporal embedding, i.e., $\mathbf{F}_{-1}^t = \mathbf{V}_{0}^t$.

After obtaining the temporal feature embeddings, we fuse the spatial feature embeddings and temporal feature embedding for the next block with: 
\begin{equation}
\mathbf{V}_{i+1} = {\rm Visual\_B}(\mathbf{V}_{i})  + \alpha \cdot \mathbf{F}^t,
\end{equation}
where $ \alpha $ is a scaling factor that controls the incorporation ratio of temporal features.

With this recurrent temporal modeling adapter, the temporal information located among the visual frames is well captured, which in turn, enrich the frame-level feature representations. 

\subsubsection{Textual Corpus Enhanced Training.}
As the limited visual samples hardly provided sufficient information for category learning, prior work~\cite{maclip} has attempted to incorporate class-level semantic information into feature learning. In this work, we propose to integrate the detailed textual descriptions into the prototype calibration to obtain a more discriminative prototype. 

To achieve this, we build a textual corpus based on the class labels with the assistance of the Large Language Models (LLMs). Specifically, given a prompt about the specific class that requests to generate some descriptions related to the corresponding class, the off-the-shelf LLMs provide a corpus containing detailed descriptions of the classes. For temporally abstract datasets like SSv2, the extended corpus maintains highly abstract textual representations, where specific objects in the corpus are replaced with generalized terms like ``object".
This enriched corpus expands the original weak textual representation space and enables a more nuanced learning process for the video encoder. 

With the class-related descriptions, we obtain the textual embeddings with the existing textual encoder, which are then used to enrich the visual representations as follows:
\begin{equation}
\mathbf{f}_{n} = \mathbf{f}_{n}^{v} + {\rm Textenc}(\text{Corpus}(class_{n})),  
\end{equation}
where $\rm Textenc$ denotes the textual encoder and we use CLIP \cite{clip} in the experiment. $\mathbf{f}_{n}^{v}$ denotes the visual feature representation of class $n$. The textual descriptions in the $n$-th corpus are concatenated as a sentence to the textual encoder. The augmented feature representations are then fed into a simple prototype merging module that is achieved with a transformer block.

\subsection{Model Fine-tuning}
\subsubsection{Unbalanced Optimal Transport Matching}
Optimal Transport (OT) \cite{cmot} employs a Sinkhorn matching algorithm to address the problem of imbalance between video content matching and temporal order matching. Considering that prototype and query videos usually lie under deviational distributions, We introduce unbalanced optimal transport (UOT) to reduce the interference of background noise from redundant frames in the matching process.

Given the support video features $\mathbf{S}$ and the query video features $\mathbf{Q}$, the distance matrix is defined as $D = [d(\mathbf{s}_i, \mathbf{q}_j)] \in \mathbb{R}^{N \times N}$, where the element $d(\mathbf{s}_i, \mathbf{q}_j)$ represents the Euclidean distance between the latent representation of the $i$-th query frame feature and that of the $j$-th support frame feature, then the similarity between query and support is denoted as:
\begin{equation}
\begin{split}
& {\rm Dis(\mathbf{Q}, \mathbf{S})}= \min_{T\in \Pi(\alpha, \beta)}\underbrace{ \langle D, T \rangle}_{OT problem} + \underbrace{\lambda \mathcal{H}(T)}_{Reg.}\\
\end{split}
\label{formula_ot}
\end{equation}
Here ${\lambda \mathcal{H}(T)}$ denotes an entropic regularization term that smooths the optimization of matching plan $T$. We set the constraints $(T \mathbf{1}) = \alpha$, ~$(T^\top \mathbf{1}) = \beta$, which naturally penalize the many-to-one matching in optimal transport. The optimal solution $T$ is called the optimal transport plan, which is the optimal distribution of latent representation pairs that minimizes the expectation of the distance $d(\mathbf{S}, \mathbf{Q})$.

Motivated by the fact that video features can contain background noise, leading to the frames between videos not necessarily requiring one-to-one correspondence, we draw inspiration from unbalanced optimal transport \cite{zhan2021unbalanced}. Originally, the score between videos was obtained through matching, as follows:
\begin{align} \label{eq: uot}
Dis(\mathbf{Q}, \mathbf{S}) &= \min_{T \in \Pi(\alpha, \beta)} \underbrace{\langle D, T \rangle}_{\text{OT problem}} + \lambda \underbrace{\mathcal{H}(T)}_{\text{Reg.}} \\  \nonumber
& + \underbrace{\tau \operatorname{KL}(T \mathbf{1}\| \mathbf{N}) + \tau \operatorname{KL}(T^{\top} \mathbf{1} \| \mathbf{N})}_{Soft Cons.}
\end{align}
where $\operatorname{KL}$ is the Kullback-Leibler divergence. And $N$ denotes the correlation distributions generated with the cross-inner product between support and query features. To obtain $T$, we followed ~\cite{zhan2021unbalanced} to approximate the UOT solution by applying the skinhorn algorithm.

\subsubsection{Training Objective}
After obtaining the prototype and query features, the metric distance between the query features and each prototype is calculated as:
\begin{equation}
Dis(\mathbf{q}, \mathbf{s}_{i}) = M(\tilde{f_{q}}, f_{i}^{s})
\end{equation}

The probability distribution of query video based on prototypes is given by:
\begin{equation}
p_{\hat{y} = i \mid q}^{fsl} = \frac{\exp(Dis(\mathbf{q}, \mathbf{s}_{i}))}{\sum_{j=1}^{N} \exp(Dis(\mathbf{q}, \mathbf{s}_{j}))}
\end{equation}
where $\hat{y}$ denotes the prediction class of the query video.

Following former works~\cite{clipfsar, maclip}, we integrate the zero-shot prediction $\textstyle p_{\hat{y} = i \mid q}^{zsl} $ with the few-shot prediction result  to get the merged prediction of query videos:
\begin{equation}
p_{\hat{y} = i \mid q} = (p_{\hat{y} = i \mid q}^{fsl})^{\lambda} \cdot (p_{\hat{y} = i \mid q}^{zsl})^{1-\lambda}
\end{equation}

The loss function consists of two parts, zero-shot video-text loss that directly predicts the action based on few-shot label and textual feature, and few-shot loss which applies cross-entropy on few-shot probability distribution:
\begin{equation}
\mathcal{L} = CE(p_{\hat{y} = i \mid q}^{fsl}, gt_{fsl}) + CE(p_{\hat{y} = i \mid q}^{zsl}, gt_{zsl}).
\end{equation}

\begin{table*}[!t]
  \centering
  \resizebox{1.0\textwidth}{!}{
        \begin{tabular}{lccccccccccc}
      \toprule
      \multirow{2}{4em}{Method} &\multirow{2}{*}{Pre-training} &\multicolumn{2}{c}{SSv2-Small}&\multicolumn{2}{c}{SSv2-Full}&\multicolumn{2}{c}{HMDB51}&\multicolumn{2}{c}{UCF101}&\multicolumn{2}{c}{Kinetics}\\ 
      & &1-shot & 5-shot & 1-shot & 5-shot & 1-shot & 5-shot & 1-shot & 5-shot & 1-shot & 5-shot\\
      \midrule
      CMN~\cite{zhu2018compound} &  INet-RN50&34.4 & 43.8 & 36.2 & 48.9&- & - & - & -&60.5 & 78.9\\ 
      OTAM~\cite{cao2020few} &  INet-RN50& -  & - & 42.8 & 52.3 &- & - & - & -&73.0 & 85.8 \\  
      AmeFuNet~\cite{fu2020depth} &  INet-RN50& - & - & - & - & 60.2 & 75.5 & 85.1 & 95.5 & 74.1 & 86.8\\
      TRX~\cite{perrett2021temporal} & INet-RN50 & - & 59.1 & - & 64.6 & - & 75.6 & - & 96.1&63.6 & 85.9 \\  
      SPRN~\cite{wang2021semantic} &  INet-RN50& - & - & - &- & 61.6 & 76.2 & 86.5 & 95.8 & 75.2& 87.1 \\
      HyRSM~\cite{wang2022hybrid} &  INet-RN50 &40.6 & 56.1 & 54.3 & 69.0 &60.3 & 76.0 & 83.9 & 94.7 &73.7 & 86.1 \\ 
      T${\rm A}^2$N + Sampler &  INet-RN50& - & - & 47.1 & 61.6 & 59.9 & 73.5 & 83.5 & 96.0 & 73.6& 86.2\\
      MoLo~\cite{wang2023molo} &  INet-RN50 &41.9 & 56.2 & 55.0 & 69.6  &60.8 & 77.4 & 86.0 & 95.5 &74.0 & 85.6 \\ 
      MGCSM~\cite{yu2023multi} &  INet-RN50& - & - &- &- & 61.3 & 79.3 & 86.5 & 97.1 & 74.2 & 88.2\\
      SA-CT~\cite{zhang2023importance} &  INet-RN50& - & - & 48.9 & 69.1 & 60.4 & 78.3 & 85.4 & 96.4 & 71.9 & 87.1 \\
      CLIP-FSAR~\cite{clipfsar} & CLIP-RN50& 52.1 & 55.8 & 58.7 & 62.8 & 69.4 & 80.7 & 92.4 & 97.0 & 90.1 & 92.0 \\ 
      \rowcolor{gray!20} \textbf{TSAM (Ours)} &  CLIP-RN50 & \textbf{53.1} & \textbf{58.0} & \textbf{60.2} & \textbf{64.3} & \textbf{72.9} & \textbf{82.8} &  \textbf{93.5} &\textbf{97.5} & \textbf{93.0} & \textbf{93.5}\\
      \midrule
      SA-CT(ViT)~\cite{zhang2023importance} & INet-ViT-B & - & - & - & 66.3 & - & 81.6 & - & 98.0 & - & 91.2 \\
      CLIP-FSAR~\cite{clipfsar} & CLIP-ViT-B &54.6 & 61.8 & 62.1 & 72.1&77.1 & 87.7 & \underline{97.0} & \underline{99.1}& 94.8 & 95.4\\ 
      MA-CLIP~\cite{maclip}  & CLIP-ViT-B &\underline{59.1} & \underline{64.5} & \underline{63.3} & \underline{72.3}&\underline{83.4} & \underline{87.9} & 96.5 &98.6& {95.7} & \underline{96.0} \\
      \rowcolor{gray!20} \textbf{TSAM (Ours)} &CLIP-ViT-B & \textbf{60.5 } & {\textbf{66.7}} & \textbf{65.8} & {\textbf{74.6}} & \textbf{84.5} & \textbf{88.9} &  \textbf{98.3} &\textbf{99.3} & \textbf{96.2} & \textbf{97.1}\\
      \bottomrule
      \end{tabular}   
  }
    \caption{Comparison results of the proposed method and the existing FSAR competitors on both 5-way 1-shot and 5-way 5-shot tasks on five benchmarks. The best results are highlighted in bold and the second-best results are underlined. The results of the competitors are from the original literature.}
  \label{tab:sota}
\end{table*}

\section{Experiments} \label{sec:exp}

\subsection{Experimental Details}
\noindent \textbf{Datasets.} We evaluate our method on four widely used few-shot action recognition datasets: HMDB-51~\cite{kuehne2011hmdb}, UCF101~\cite{soomro2012ucf101}, Kinetics~\cite{carreira2017quo}, and SSv2~\cite{goyal2017something}. The first three datasets focus on scene understanding, while SSv2 requires more challenging temporal modeling~\cite{Tu_2023_ICCV}. For both HMDB-51 and UCF101, we use the few-shot split from \cite{zhang2020few}, with 31/10/10 and 70/10/21 train/val/test classes, respectively. Following~\cite{zhu2018compound}, we select a subset of Kinetics with 64/12/24 train/val/test classes. For SSv2, we report results on two few-shot splits: SSv2-Small~\cite{zhu2020label} and SSv2-Full~\cite{cao2020few}. Both have 64/12/24 train/val/test classes, with SSv2-Full containing 10$\times$ more videos per class in the training set.

\noindent \textbf{Implementation Details.} We adapt two CLIP pre-trained \cite{clip} backbones, ResNet50 and ViT-B/16, for comparison. Each video is uniformly sampled into 8 frames, which are then cropped to a resolution of 224$\times$224. During training, we freeze the pre-trained backbone and fine-tune only the proposed adapter. We use SGD as the optimizer. For testing, we sample 10,000 few-shot action tasks to calculate the 5-way 1-shot and 5-way 5-shot average accuracy. Training is conducted on 4 RTX 3090 GPUs.

\subsection{Performance Comparison}
\cref{tab:sota} provides a comprehensive comparison of our method and 12 recent few-shot action recognition competitors on five datasets. The competitors use either CLIP or the ResNet model pre-trained on the ImageNet-21k as the visual backbone. From the results, we observe that our TSAM achieves the best performances in both 1-shot and 5-shot tasks. Specifically, for the 1-shot scenario, our method surpasses the second-best competitor by significant margins of 1.4\%, 2.5\%, 1.1\%, 1.3\%, and 0.5\% on SSv2-Small, SSv2-Full, HMDB-51, UCF101, and Kinetics datasets, respectively. In the 5-shot setting, our approach respectively outperforms the second-best competitors by 2.2\%, 2.3\%, 1.0\%, 0.2\%, and 1.1\% on each of these datasets, highlighting its effectiveness.

Besides, we also reach the following observations: \textbf{1)} Backbone matters: the visual backbones trained with the Transformer-based architecture (i.e., ViT-B) demonstrate notably superior performance in comparison to those utilizing the ResNet-based architecture (i.e., ResNet50), highlighting the significance of the pre-trained visual backbone as a crucial factor contributing to video recognition capabilities. \textbf{2)} Fine-tuning manner: Compared to CLIP-FSAR \cite{clipfsar}, which fully fine-tunes the backbone, our temporal adapter structure offers a more nuanced representation of temporal information while requiring fewer parameters, effectively balancing the pre-trained model's capabilities with video feature adaptation. \textbf{3)} Temporal modeling: Compared to other adapter-based methods, \textit{e.g.}, MA-CLIP \cite{maclip}, our sequential perceiver structure injects sequential information into the pre-trained backbone, enabling a stronger video encoder that constructs a more aligned video feature space for support and query videos. For example, TSAM outperforms MA-CLIP \cite{maclip} by 2.2\% in the 1-shot on the SSv2-Full dataset, which demands comprehensive temporal understanding.

\begin{figure}[!t]
    \centering
    \includegraphics[width=1.0\linewidth]{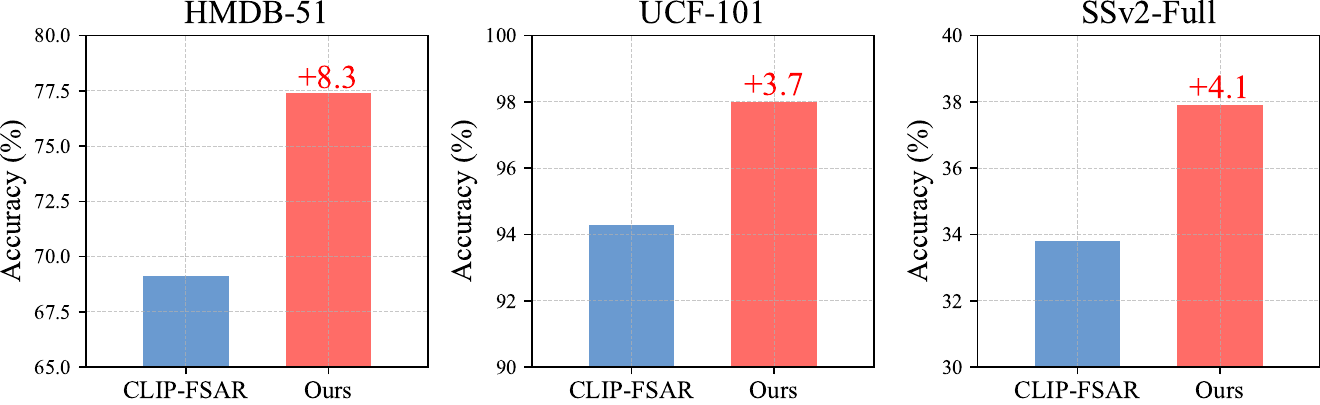}
\caption{Comparison result of our method and the competitor CLIP-FSAR \cite{clipfsar} under cross-dataset evaluation.}
\label{fig:cross-dataset}
\end{figure}

\noindent \textbf{Cross-Dataset Comparison.}
In this experiment, we embark on a cross-dataset evaluation to assess the model's generalization capabilities. Specifically, we train the model on the Kinetics dataset and then evaluate the model on UCF-101, HMDB-51, and SSv2 datasets. As depicted in \cref{fig:cross-dataset}, our method exhibits a notable advantage over CLIP-FSAR \cite{clipfsar}, achieving impressive improvements of 3.7\% on UCF-101, 8.1\% on HMDB-51, and a substantial 3.9\% gain on the challenging SSv2 dataset. This consistent superior performance across diverse video datasets underscores the robust generalization capabilities of our proposed approach, effectively excelling in varied video data distributions.

\subsection{Ablation Studies}
\subsubsection{Study on Partial Adaption.}
\begin{figure}[!tb]
    \centering
     \setlength{\tabcolsep}{4pt}
     \includegraphics[width=0.98\linewidth]{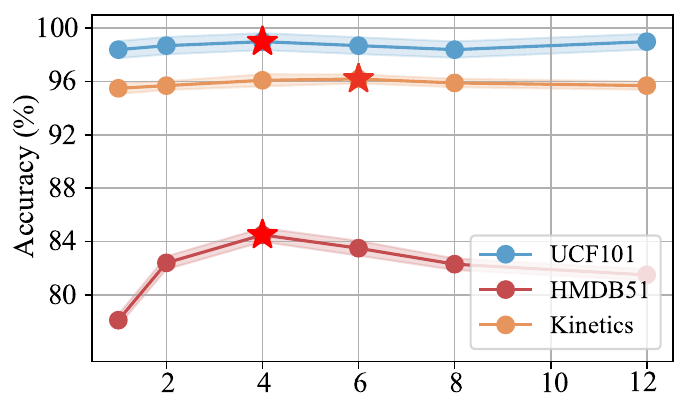}
     \vspace{-0.1cm}
    \caption{Effect of different partial injection numbers}
    \label{fig:ft-methods}
\end{figure}

In this experiment, we evaluate the impacts of adapter numbers added during the model fine-tuning.  \cref{fig:ft-methods} describes the results vary with the number of adapters added on the three datasets. When dealing with datasets such as UCF101 and Kinetics100, which inherently exhibit high accuracy levels, the addition of more adapters results in only marginal improvements in performance. In contrast, for the HMDB-51 dataset, adding a suitable number of adapters leads to significant gains. Specifically, four adapters are shown to be sufficient to achieve a substantial performance boost, effectively reducing the number of parameters that need to be learned and significantly lowering the computational cost during both inference and training. It is worth noting that the adapters are added from the last module backward, as the later modules in the pre-trained backbone extract more abstract semantic information during video frame encoding, making them more efficient for temporal information extraction.

\begin{table}[!t]
  \centering
 \resizebox{0.95\linewidth}{!}{
 \tiny
 \begin{tabular}{ccc|cc} 
 \toprule
 \textbf{S.P} & \textbf{T.C} & \textbf{UOTM} & \textbf{SSv2-Small} & \textbf{HMDB-51} \\
 \midrule
 \xmark   &  \xmark    &  \xmark    & 54.6       & 77.1 \\
 \cmark  &  \xmark    &  \xmark   & 58.2       & 82.2 \\
  \cmark  &  \cmark  &  \xmark    & 59.0       & 83.9 \\
  \cmark  &  \xmark    &  \cmark  & 58.7       & 82.6 \\
 \midrule
  \cmark  &  \cmark  &  \cmark  & \textbf{60.2} & \textbf{84.5}     \\
 \bottomrule
 \end{tabular}
 }
\caption{Effectiveness (\%) of each component. We compare the performance of different configurations for the 5-way 1-shot task on the SSv2-Small and HMDB-51 datasets.}
\label{tab:ab-components}
\end{table}

\subsubsection{Impacts of Each Component}
In \cref{tab:ab-components}, we investigate the role of each designed module in the TSAM framework on the two datasets. The results show the performance improvements as each module is incrementally added to the baseline model. The baseline configuration, without any of the designed modules (S.P, T.C, or S.N), achieves an accuracy of 54.6\% on SSv2-Small and 77.1\% on HMDB-51, respectively. When the Sequential Perceiver (S.P) module is incorporated, a noticeable improvement is observed, with the accuracy rising to 58.2\% on SSv2-Small and 82.2\% on HMDB-51, respectively. Adding the Textual Corpus (T.C) module further improves performance, boosting accuracy to 59.0\% on SSv2-Small and 83.9\% on HMDB-51, respectively. Similarly, incorporating the Unbalanced Optimal Transport Matching (UOTM) module alongside S.P also results in better performance, with an accuracy of 58.7\% on SSv2-Small and 82.6\% on HMDB-51. The S.N module helps normalize and harmonize the feature representations, ensuring consistency across different video sequences. Finally, when all three modules (S.P, T.C, and S.N) are combined, the model achieves the highest performance, with an accuracy of 60.2\% on SSv2-Small and 84.5\% on HMDB-51, respectively. This demonstrates the complementary nature of these modules, with each contributing uniquely to the overall effectiveness of the TSAM framework.

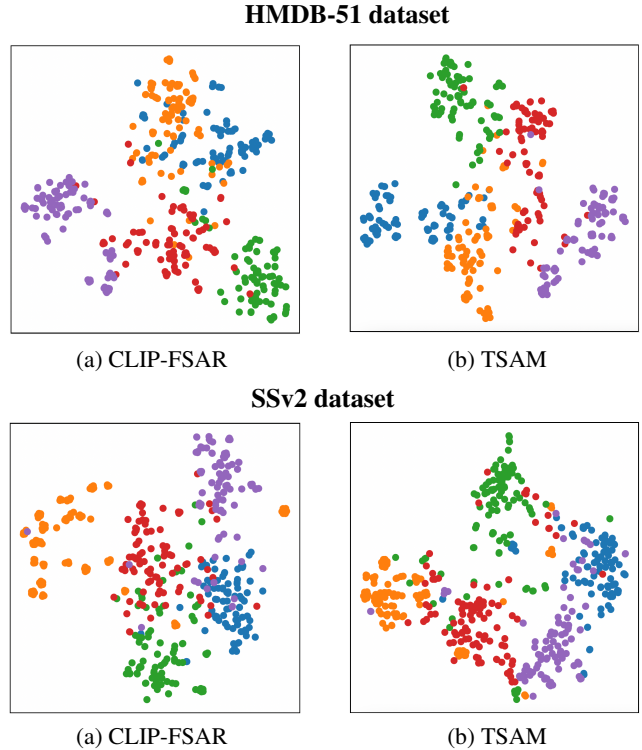
\begin{figure}[t]
   \begin{overpic}[width=\linewidth]{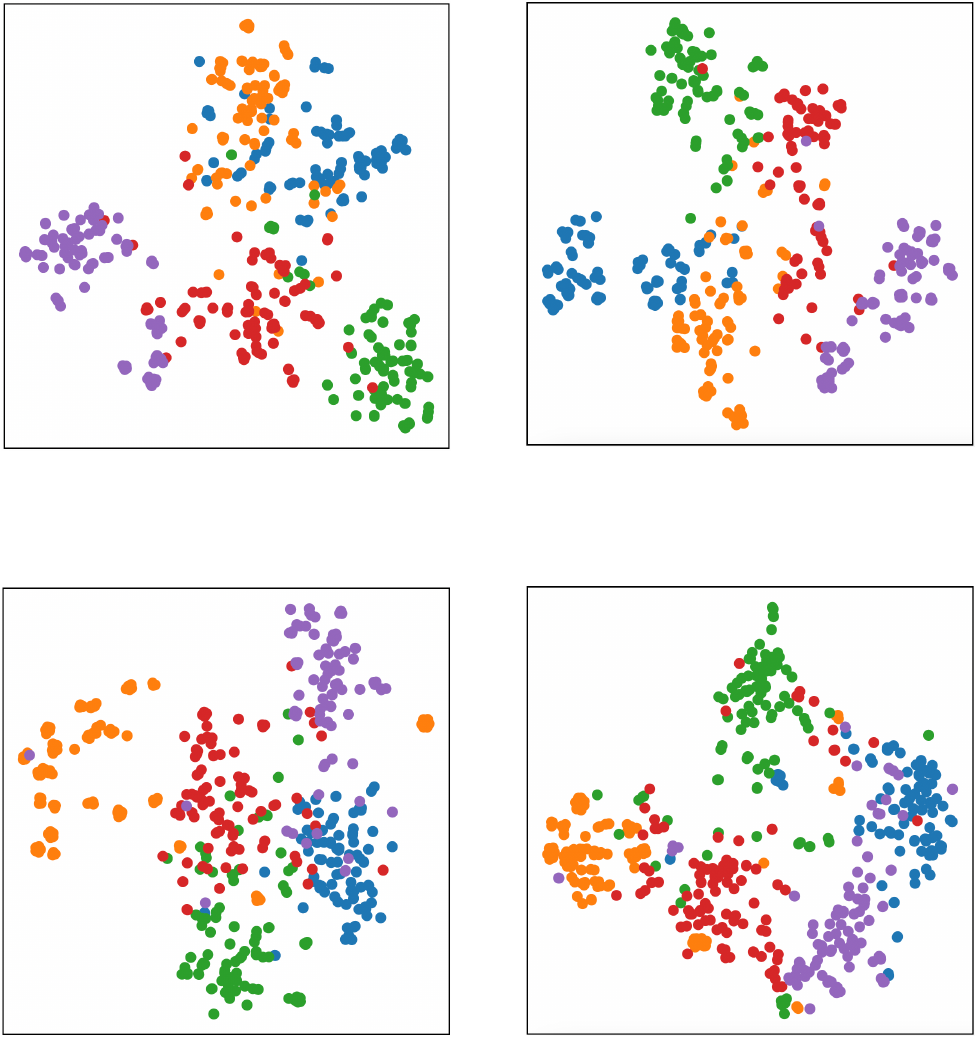} 
   \put(10,52){\small (a) CLIP-FSAR}
   \put(65,52){\small (b) TSAM}
   \put(34,103){ \textbf{HMDB-51 dataset}}
   \put(35,46){ \textbf{SSv2 dataset}}
    \put(10,-4){\small (a) CLIP-FSAR}
   \put(65,-4){\small (b) TSAM}
   \end{overpic}
    \vspace{5mm}
   \caption{t-SNE visualization of the competitor CLIP-FSAR \cite{clipfsar} and our TSAM on both HMDB-51 and  SSv2 datasets. For HMDB51 dataset, five classes are \textcolor{RoyalBlue}{'fencing'}, \textcolor{orange}{'kick'}, \textcolor{OliveGreen}{'kick ball'}, \textcolor{BrickRed}{'pick'} and \textcolor{Orchid}{'pour'}. For SSv2 dataset, five classes are \textcolor{RoyalBlue}{'Approaching [something] with your camera'}, \textcolor{orange}{'Poking a stack of [something] without the stack collapsing'},\textcolor{OliveGreen}{'Pretending to open [something] without actually opening it'}, \textcolor{BrickRed}{'pushing something from right to left'} and \textcolor{Orchid}{'showing [something] next to something'}.}
   \label{fig:tsne}
\end{figure}

\subsection{Further Analysis}
\subsubsection{Temporal Reverse Comparison.}
In this experiment, we assess the temporal modeling capability of the models by subjecting them to both standard and temporally reversed video inputs for a comparative analysis. Intuitively, model performance should decline significantly when the input videos are reversed, as the actions and temporal relationships are disrupted. While many existing methods claim robust temporal modeling abilities, they often fail to exhibit this robustness under such conditions, as shown in \cref{tab:temporalreverse}. The results indicate that TSAM experiences notable performance drops, \textit{e.g.}, 3.2\%, 1.1\%, and 0.8\% on the SSv2-Full, HMDB-51, and Kinetics-100 datasets, respectively. Notably, on datasets like SSv2, which heavily rely on temporal information, TSAM shows a significant decline of 3.2\%. This suggests that our design effectively captures sequence-dependent temporal information rather than merely relying on recognizing specific objects within frames. For example, on HMDB-51 and UCF-101 datasets, the performance of the competitors remains relatively stable, failing to demonstrate a notable drop when subjected to temporal reversals.

\definecolor{tabcolor}{RGB}{200, 200, 200}
\def\tabwidth{.31}
\begin{table}[!t]
    \centering
    \resizebox{0.9\linewidth}{!}{
            \begin{tabular}{l|cc}
                \toprule
                \textbf{Dataset} & \textbf{CLIP-FSAR} & \textbf{TSAM} \\
                \midrule
                SSv2-Full & 62.1 & 65.5 \\
                SSv2-Full-reverse & 62.0 & 62.3 (↓3.2\%) \\
                \midrule   
                HMDB-51 & 77.1 & 84.5 \\
                HMDB-51-reverse & 77.1 & 83.4 (↓1.1\%) \\
                \midrule
                Kinetics100 & 94.7 & 96.2 \\
                Kinetics100-reverse & 94.8 & 95.4 (↓0.8\%) \\
                \bottomrule
            \end{tabular}
        }
        \caption{Temporal Reverse comparison between CLIP-FSAR and our TSAM on three datasets.}
        \label{tab:temporalreverse}
\end{table}

\subsubsection{t-SNE Visualization}
We visualize the feature distribution changes of CLIP-FSAR and our TSAM during the test stage, as illustrated in \cref{fig:tsne}. Five classes are randomly sampled from the test sets of HMDB-51 and SSv2 for this visualization. By comparing the results, we observe a clear improvement with our method, which yields more compact intra-class distributions and more discriminative inter-class features. Notably, on the challenging SSv2 dataset, our TSAM demonstrates superior performance, resulting in more distinguishable classes and tighter intra-class distributions.

\section{Conclusion} \label{sec:conclusion}
In this paper, we have proposed a novel FSAR framework named Temporal Sequential-Aware Model (TSAM), utilizing sequential perceiver as an adapter to transfer the pre-trained image backbone to the video encoder. To further improve the few-shot matching process, we expanded the textual corpus to construct multimodal prototypes and unbalanced optimal transport matching metrics to filter extra background noise within video frames. Our framework emphasizes temporal order during the video encoding process and metric matching, greatly outperforms the former method across all FSAR benchmarks, leading to 2.5\% improvement on the temporal-challenging SSv2 datasets.
\bibliography{aaai25}

\end{document}